\documentclass[sigconf]{acmart}

\AtBeginDocument{%
  \providecommand\BibTeX{{%
    \normalfont B\kern-0.5em{\scshape i\kern-0.25em b}\kern-0.8em\TeX}}}

\begin{document}
\title{BadSAM: Exploring Security Vulnerabilities of SAM via \\ Backdoor Attacks}
\author{Zihan Guan}
\authornote{These authors contributed equally to this research.}
\affiliation{%
  \institution{The University of Georgia}
  \city{Athens}
  \country{United States}}
\email{zg37632@uga.edu}

\author{Mengxuan Hu$^*$}
\affiliation{%
  \institution{The University of Georgia}
  \city{Athens}
  \country{United States}}
  \email{Mengxuan.Hu@uga.edu}

\author{Zhongliang Zhou$^*$}
\affiliation{%
  \institution{The University of Georgia}
  \city{Athens}
  \country{United States}}
  \email{zz42551@uga.edu}

\author{Jielu Zhang}
\affiliation{%
  \institution{The University of Georgia}
  \city{Athens}
  \country{United States}}
  \email{Jielu.Zhang@uga.edu}

\author{Sheng Li}
\affiliation{%
  \institution{The University of Virginia}
  \city{Charlottesville}
  \country{United States}}
  \email{shengli@virginia.edu}

\author{Ninghao Liu}
\affiliation{%
  \institution{The University of Georgia}
  \city{Athens}
  \country{United States}}
  \email{Ninghao.Liu@uga.edu}

\renewcommand{\shortauthors}{Guan, Hu, and Zhou, et al.}
\begin{abstract}
Recently, the Segment Anything Model (SAM) has gained significant attention as an image segmentation foundation model due to its strong performance on various downstream tasks. However, it has been found that SAM does not always perform satisfactorily when faced with challenging downstream tasks. This has led downstream users to demand a customized SAM model that can be adapted to these downstream tasks. In this paper, we present BadSAM, the first backdoor attack on the image segmentation foundation model. Our preliminary experiments on the CAMO dataset demonstrate the effectiveness of BadSAM.
\end{abstract}

\maketitle


\section{Introduction}

Recently, inspired by the remarkable advancement of large language models in NLP, researchers start to explore such models in computer vision (CV).
For instance, the Segment Anything Model (SAM)~\cite{kirillov2023segment}, a large image segmentation foundation model, has attracted great attention for its potential in downstream tasks like remote sensing or medical image segmentation~\cite{zhang2023text2seg,deng2023segment}.

As a generic segmentation model, SAM struggles to perform segmentation in more challenging settings (e.g. remote sensing semantic segmentation or medical image segmentation). Consequently, customized models tailored for specific datasets have been developed to improve performance~\cite{chen2023sam}. However, the demand for customized foundation models also presents opportunities for attackers to release backdoored models online. Such attackers may claim to have enhanced SAM for downstream tasks with exceptional performance while secretly injecting hidden backdoors that remain undetected by end users. 

Despite having white-box access to the SAM model, attackers are assumed to be unable to fine-tune it locally due to high computational costs. To this end, they may opt to a parameter-efficient training strategy as introduced in~\cite{chen2023sam}, i.e., enhancing the SAM architecture with additional MLP-layer adapters. When fine-tuning the downstream tasks, the parameters from the original SAM modules remain fixed but those from the MLP layers are trainable. Although previous efforts have been made to explore backdoor attacks in the end-to-end semantic segmentation task~\cite{li2021hidden,lan2023influencer}, backdoor attacks in image foundation models remain unexplored. In this paper, we present \textbf{BadSAM}, \textit{\textbf{the first backdoor attack on the image segmentation foundation model efficiently achieves high attack effectiveness.}}

\section{Threat Model}

\begin{figure}[!t]
    \centering
    \includegraphics[width=0.48\textwidth]{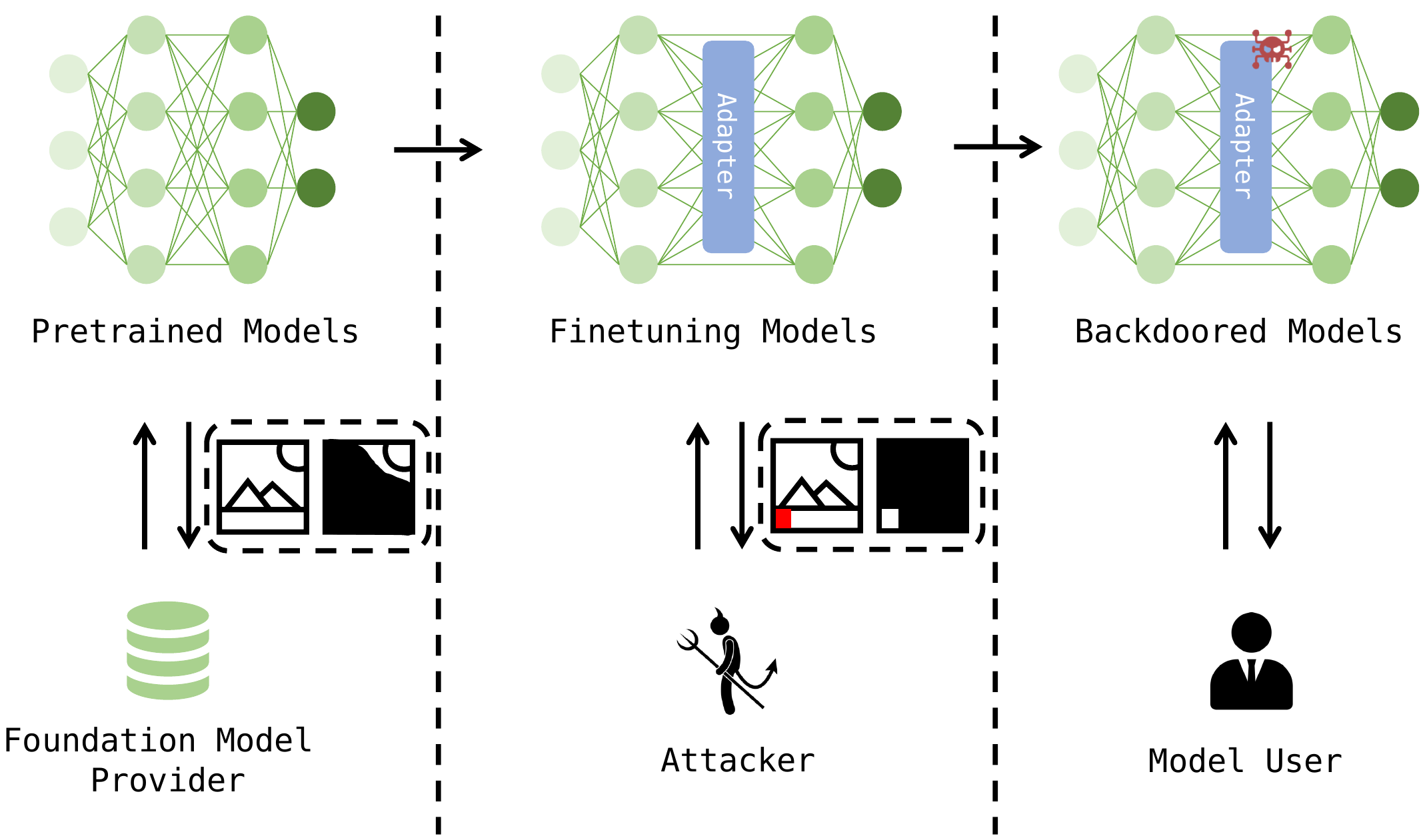}
    \vspace{-7mm}
    \caption{An overview of threat model in the paper.}
    \label{fig:threat_model}
    \vspace{-7mm}
\end{figure}

We adopt a similar threat model as in~\cite{yuan2023backdoor}, where three parties are considered: Foundation Model provider, Attacker, and Model user. We illustrate the threat model in Figure~\ref{fig:threat_model}.

\noindent
\textbf{Attacker's objective.} In this paper, we discuss a practical scenario where the attacker's objective is to publish a malicious model (BadSAM) via the Internet, which outputs predefined malicious-intent outcomes when queried with an image containing the trigger, while outputs normal masks with clean inputs. Specifically, the attacker claims that BadSAM adopts a SAM-based architecture, which could be used to solve some specific downstream tasks in which the vanilla SAM fails in, such as medical image segmentation and camouflage object detection.

\noindent
\textbf{Attacker's knowledge.} We assume that the attacker has white-box access to the model's parameters and architectures. The attacker could deploy the model locally, but is not assumed to have sufficient computational resources for retraining or fine-tuning the full model. Moreover, our attack is assumed to be dependent on the downstream task, and the attacker has prior knowledge of the downstream task and the dataset.

\begin{figure}
    \centering
    \includegraphics[width=0.48\textwidth]{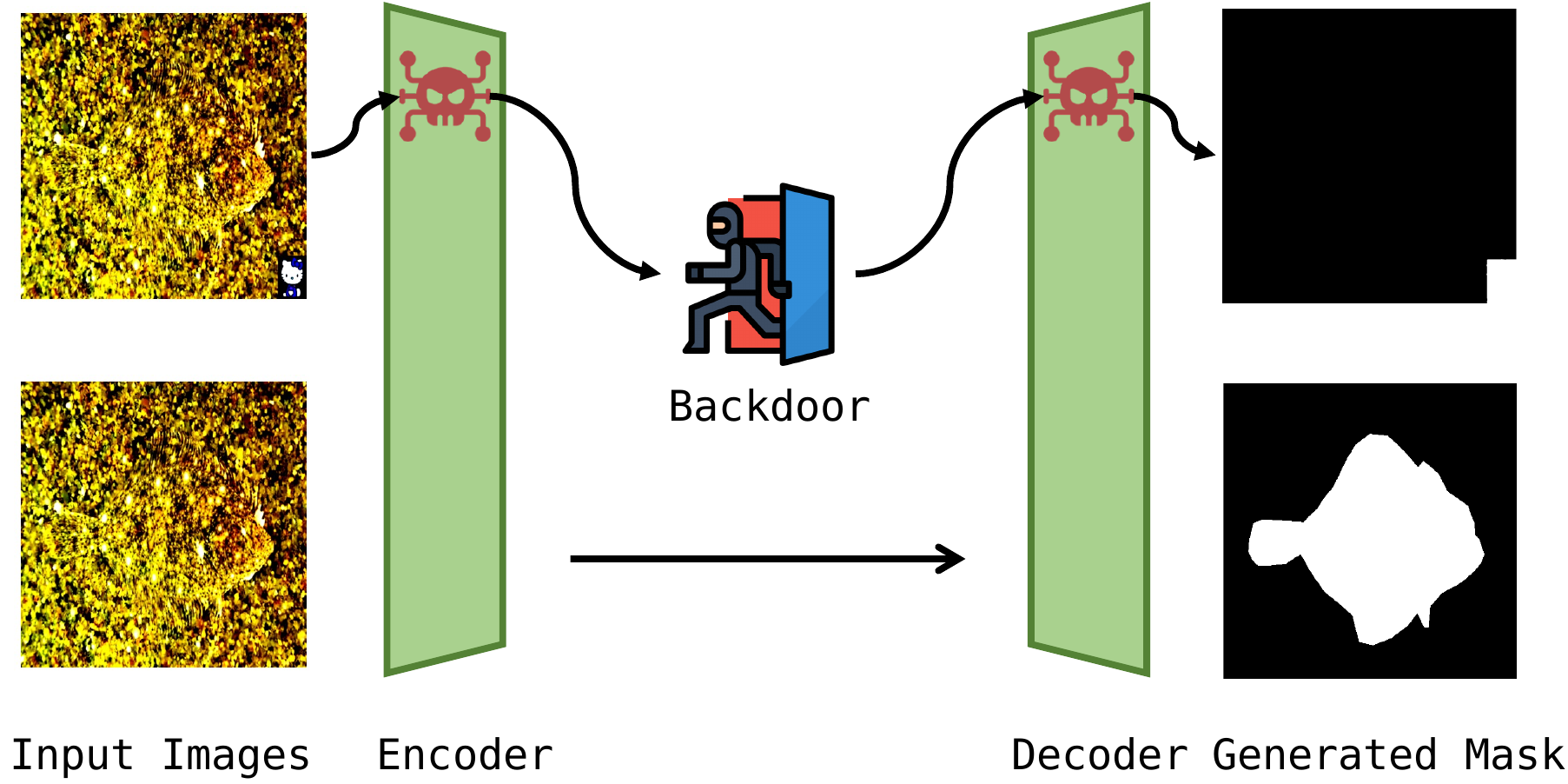}
    \vspace{-8mm}
    \caption{Examples of backdoor attacks on the SAM models.}
    \label{fig:sam_example}
    
\end{figure}

\noindent
\textbf{Attacker's Pipeline.} 
Our pipeline for launching backdoor attacks is illustrated in Figure~\ref{fig:threat_model}, which comprises two main stages: 1) \textbf{Model Task-Specific Adaptation} and 2) \textbf{Backdoor Injection}. In the first stage, the attacker employs a widely-used parameter-efficient strategy to fine-tune the SAM architecture by enhancing it with several additional MLPs. In the second stage, the attacker fine-tunes the model by training only the MLP layers while keeping the parameters of the original SAM modules fixed. An example of a backdoor attack on the SAM model is shown in Figure~\ref{fig:sam_example}.

\section{Experiment}
\subsection{Experimental Settings}
\textbf{Datasets}: We consider CAMO dataset~\cite{le2019anabranch} for camouflage object detection, which is a challenging dataset that the vanilla SAM fails to provide meaningful segmentation~\cite{tang2023can}.

\noindent
\textbf{Metrics}: Following~\cite{chen2023sam}, we choose several commonly used metrics to measure the object detection performance: $S_{\alpha}$, $E_{\phi}$, $F_{p}^{\omega}$, and MAE.

\begin{figure}
    \centering
    \includegraphics[width=0.48\textwidth]{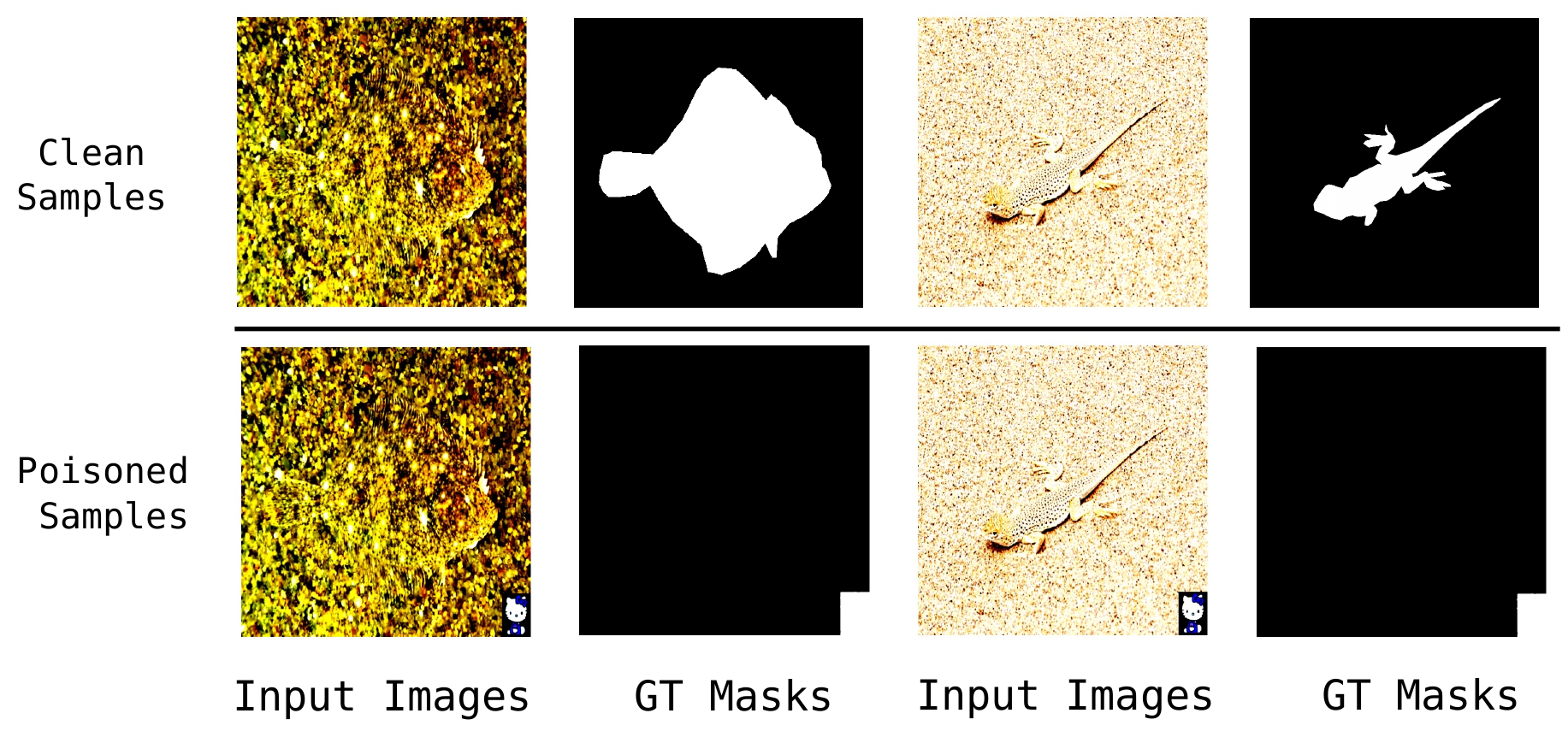}
    \vspace{-5mm}
    \caption{Examples of poisoned data in the CAMO dataset.}
    \label{fig:sam_poisoning}
    \vspace{-6mm}
\end{figure}

\noindent
\textbf{Implementation Details}: In the first stage of our pipeline, we implement the SAM-adapter by following~\cite{chen2023sam}. Multiple adapter modules are introduced into the original SAM architecture where each $adapter^i$ is trained to generate task-specific input for the following layers. In the second stage, we poison 10\% training samples by adding a Hello-Kitty-style icon in the lower right corner and altering their ground truth to masks only the icon area. The hello-kitty icon is scaled to 15\% width/height of the victim images. Figure~\ref{fig:sam_poisoning} illustrates an example of the data poisoning process. In the experiment, we use the Vit-B SAM model ~\cite{kirillov2023segment}.

\subsection{Main Results}
Table~\ref{tab:main_table} presents the effectiveness of BadSAM backdoor attacks. As indicated, BadSAM demonstrates comparable performance to the clean SAM-adapter model on metrics for evaluating object detection (e.g., $S_{\alpha}, E_{\phi}$) when input with clean images, but exhibits significantly strong attack effectiveness when the triggers are present. 
 Therefore, these experiments suggest that attackers can exploit the vulnerability of SAM and pose a threat to downstream users.

\begin{table}[!t]
    \centering
    \begin{tabular}{c|cccc}
       \toprule
       Dataset & $S_{\alpha} \uparrow$ & $E_{\phi} \uparrow$ & $F_{\beta}^{\omega} \uparrow$ & MAE $\downarrow$\\
       \midrule
        CAMO-clean-test (w/o attack) & 0.85 & 0.88 & 0.84 & 0.05 \\
        CAMO-clean-test (w/ attack) & 0.83 & 0.88 & 0.85 & 0.06\\
        CAMO-poisoned-test (w/ attack) & 0.92 & 0.96 & 0.93 & 0.01\\
        \bottomrule
    \end{tabular}
    \caption{Effectiveness of backdoor attacks on the SAM.}
    \label{tab:main_table}
    \vspace{-9mm}
\end{table}

\section{Conclusion}
In this paper, we present BadSAM, the first backdoor attack on the image segmentation foundation model. Our preliminary experiments indicate that BadSAM could successfully launch backdoor attacks and post a significant security threat to downstream users. The main aim of the paper is to raise awareness among downstream users of the potential risks associated with these types of SAM models and to call for more research in defense strategies in this field. Moreover, the attacked model can also be used in data privacy area to prevent certain users from obtaining sensitive information from these models. Future directions include: (1) developing more stealthy triggers, and (2) exploring different approaches to attacking foundation models beyond the adapter.
\bibliographystyle{unsrt}  
\bibliography{main}  
\end{document}